%% file: iclr2025_conference.tex
\documentclass{article} 
\usepackage{iclr2025_delta,times}

\input{math_commands.tex}

\usepackage{hyperref}
\usepackage{url}
\usepackage{graphicx}
\usepackage{svg}
\usepackage{placeins}  

\usepackage{xcolor}
\definecolor{rot}{HTML}{980C13}
\definecolor{blau}{HTML}{1F77B4}
\definecolor{gruen}{HTML}{26A259}

\title{Latent Diffusion U-Net Representations Contain Positional Embeddings and Anomalies}

\iclrfinalcopy 
\begin{document}
\author{Jonas~Loos\textsuperscript{1} \And Lorenz Linhardt\textsuperscript{1,2}}    
\maketitle
\renewcommand{\thefootnote}{\arabic{footnote}}
\footnotetext[1]{Machine Learning Group, Technische Universit\"at Berlin, Berlin, 10623, Germany}
\footnotetext[2]{BIFOLD - Berlin Institute for the Foundations of Learning and Data, Berlin, 10623, Germany}

\begin{abstract}
Diffusion models have demonstrated remarkable capabilities in synthesizing realistic images, spurring interest in using their representations for various downstream tasks. To better understand the robustness of these representations, we analyze popular Stable Diffusion models using representational similarity and norms. Our findings reveal three phenomena: (1) the presence of a learned positional embedding in intermediate representations, (2) high-similarity corner artifacts, and (3) anomalous high-norm artifacts. These findings underscore the need to further investigate the properties of diffusion model representations before considering them for downstream tasks that require robust features. Project page: \url{https://jonasloos.github.io/sd-representation-anomalies}.
\end{abstract}

\section{Introduction}
Ever since diffusion models~\citep{sohl2015deep, song2019, ho2020denoising} superseded generative adversarial networks~\citep{goodfellow2020generative} in image generation~\citep{dhariwal2021diffusion}, diffusion methodology progressed steadily. Architectural and training improvements, such as latent diffusion~\citep{rombach2022HiResSynth}, transformer-based architectures~\citep{peebles2023scalable,esser2024scaling}, and model distillation~\citep{sauer2023adversarial,sauer2024fast} allow for more efficient training and generation of higher quality images.

Improvements in efficiency, together with remarkable image generation abilities have led to investigations into image diffusion models as embedding models (e.g.~\citep{xiang2023denoising}). Similar to DINO~\citep{dino, dinov2} or CLIP~\citep{clip} models, diffusion models may yield representations useful for downstream tasks (e.g. classification~\citep{xiang2023denoising} or semantic correspondence~\citep{zhang2023tale}). 
Yet, attempts to use pretrained diffusion models as embedding models, as well as investigations of their general capabilities, have revealed limitations, such as texture bias in higher layers~\citep{zhang2023tale}, insufficient linguistic binding~\citep{rassin2023}, and left-right confusion~\citep{zhang2024telling}. We refer to Appx.~\ref{appx:rw} for additional related work.

In this work, we present three novel empirical phenomena in image diffusion model representations that do not encode spatially localized semantics and thus may deteriorate downstream task performance. We focus on the popular U-Net-based Stable Diffusion (SD) models, as they have been repeatedly investigated for their downstream utility (e.g.~\citep{zhang2023tale,zhang2024telling,tang2023emergent,baranchuk2022labelefficient,zhao2023unleashing,Ke2024RepurposingDI}).
The main contributions of this work are:
\begin{itemize}
    \item{(C1)} We show that the representations of models of the SD family encode a \textit{positional embedding}. This embedding is linearly extractable from the representations of lower blocks.
    \item{(C2)} We show that representations of lower blocks often contain corner \textit{tokens of abnormally high similarity} to other corner tokens. This phenomenon is independent of the image content and can even be observed between tokens of different images.
    \item{(C3)} We show that representations of lower blocks sometimes contain \textit{tokens of abnormally high norm} that do not appear to capture only the local image content.
\end{itemize}

\section{Methods}

\subsection{Representation Extraction}
We extract intermediate representations from the U-Net layers of the evaluated models. The architecture consists of a series of four down-sampling (\texttt{dn0}-\texttt{dn3}) and four up-sampling (\texttt{up0}-\texttt{up3}) blocks, connected by skip connections, as well as a resolution-preserving \texttt{mid}-block at the lowest level. Each block consists of a combination of ResNet and attention layers, and a down- or up-sampling operation where applicable. We extract representations after each layer by noising a given image $x$ in the latent space of the variational autoencoder $\mathcal{E}$ according to a given time step $t \in [1,1000]$ and then recording the activations after each U-Net layer. More formally:
\begin{align}
    z_0=\mathcal{E}(x), \quad z_t = \sqrt{\bar{\alpha}_t} z_0 + \sqrt{1 - \bar{\alpha}_t} \epsilon, \quad 
r_l = \text{U-Net}(z_t, t, \mathcal{C})_l,
\end{align}
where $\bar{\alpha}$ is defined by the noise scheduler and interpolates between the latent code $z_0$ of the image and the noise $\epsilon$. We set conditioning $\mathcal{C}$ to an empty prompt and $t=50$. The representation $r_l$ at a layer $l$ is of size $\mathbb{R}^{w_l \times h_l \times c_l}$, with $w_l$, $h_l$, and $c_l$ being the width, height, and number of channels, respectively. The spatial dimensionality decreases in lower layers of the U-Net, while $c_l$ increases. We refer to the representations at any spatial position as a token. 

\subsection{Position Estimation}

To quantify the observation of positional embeddings in the representations of SD models, we train linear probe to estimate the token position. An estimator for layer $l$ takes as input a token of dimensionality $c_l$ and predicts the vertical and horizontal coordinate of the token, using labels formed by concatenating two one-hot vectors of dimensionality $w_l + h_l$. We minimize the cross-entropy loss using the Adam optimizer~\citep{kingma2017adam} with learning rate $10^{-3}$ for 5 epochs.

\section{Experiments and Analysis}

\begin{figure}
\includegraphics[width=1\textwidth]{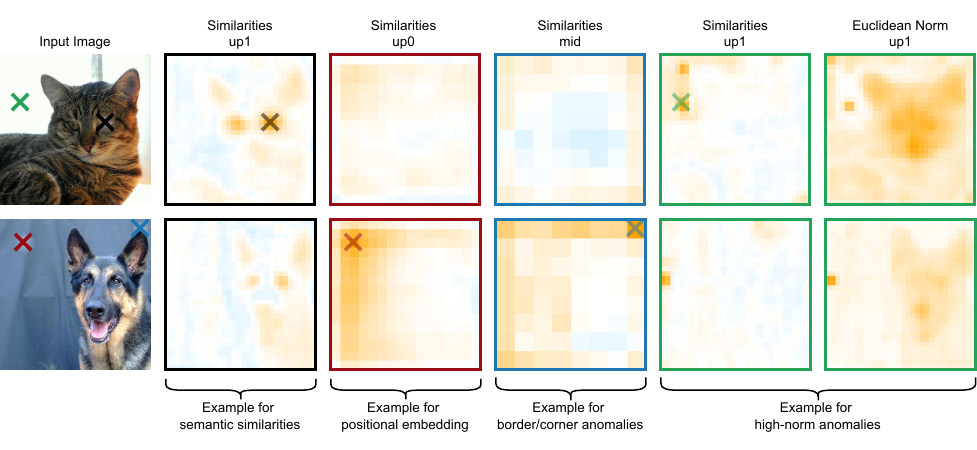}
\caption{Cosine similarity and Euclidean norm across spatial positions of representations. Each column shows an example of one of the three observations, or of meaningful similarities. Similarities in each column are relative to the token highlighted by a marker of the matching color (\textbf{×}, \textcolor{red}{\textbf{×}}, \textcolor{blue}{\textbf{×}}, \textcolor{gruen}{\textbf{×}}) in one of the images. Representations are extracted from SD-1.5 at the blocks indicated at the top.}
\label{fig:results-examples}
\end{figure}

In this section, we present our findings on positional embeddings, the influence of 
corner and border locations, and high norm anomalies. For each phenomenon, we provide a qualitative example, a quantification of the observation, and a brief discussion of potential implications.

For all experiments, we extract representations from the U-Net-based latent diffusion models SD-1.5, SD-2.1 \citep{rombach2022HiResSynth}, and SD-Turbo \citep{sauer2023adversarial}. For brevity, we show the results for the latter two in Appx.~\ref{appx:models}.
We use a subset of ImageNet \citep{ILSVRC15}, containing 100 randomly chosen images for each of 5 classes (\textit{German shepherd} (235), \textit{boxer} (242), \textit{tiger cat} (282), \textit{pickup} (717), \textit{volcano} (980)), which were chosen to contain both concepts of high as well as low similarity. All images are center-cropped and resized to $512\times 512$ pixels to match the default image size of SD-1.5.

\subsection{SD U-Net Representations Contain positional embeddings}\label{sec:spatial}
\paragraph{Qualitative observation.}
We observe that representations in SD models contain a positional embedding, which is visible in the \textcolor{rot}{third column} of Fig.~\ref{fig:results-examples}, where tokens of similar spatial locations show higher similarities, even across images. This implies that SD models saliently encode location in their representations. We find that this phenomenon is most apparent after the \texttt{up0} block and less visible in higher blocks. The following quantitative results support this observation.

\begin{figure}
\centering
\includegraphics[width=.9\textwidth]{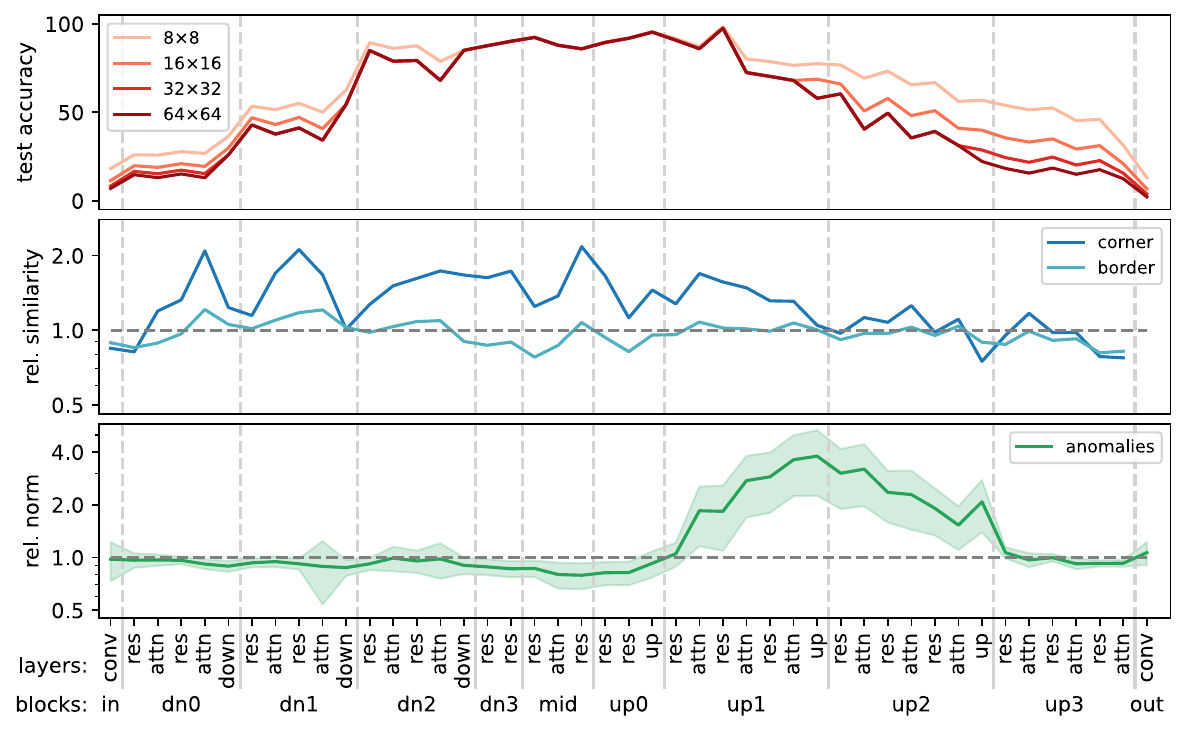}
\caption{Quantitative results for position estimation, border/corner artifacts, and high-norm anomalies for SD-1.5. \textcolor{rot}{\textbf{Top row:}} Linear probe accuracy for position estimation. Brighter shades indicate reduced resolution. \textcolor{blau}{\textbf{Middle row:}} Relative similarity of tokens lying at a border/corner of the cropped images w.r.t. their similarity before cropping. (log-2 scale). \textcolor{gruen}{\textbf{Bottom row:}} Relative average norm of anomalous tokens w.r.t. to the mean norm of all tokens of the same representation (log-2 scale). }
\label{fig:results-quantitative}
\end{figure}

\paragraph{Quantitative results.}
To quantify the positional embedding uncovered by inspecting token similarities, we train a linear probe to predict the spatial location of each token given a representation token as input. We use a random 80\% split of the images for training and evaluate on the remaining 20\% by calculating the fraction of correct width and height predictions. As shown in the \textcolor{rot}{first row} of Fig.~\ref{fig:results-quantitative}, the estimator achieves a test accuracy of over 90\% for lower blocks (\texttt{down2} to \texttt{up1}), indicating that the positional information is more saliently encoded there. Part of the difference in performance across layers is due to the lower spatial resolution of the lower blocks. Yet, even when evaluating the performance of the higher blocks at a lower resolution by coarse-graining the prediction target, lower blocks still yield significantly higher accuracy.

\paragraph{Implications.} SD models saliently encode spatial locations in their representations to generate images, which has immediate consequences for their use as representation learners. For example, in semantic or dense correspondence tasks, similarity between representation tokens is used to determine semantically matching image locations across two images. Saliently encoded position information may interfere with semantic matching, undermining task performance.

\subsection{SD U-Net Representations Contain Corner Artifacts}\label{sec:border-art}
\paragraph{Qualitative observation.}
We find that tokens located at the corners and borders often have unusually high cosine similarities to each other, even if there is no obvious correspondence of the image content. This anomalous behavior of the border and corner tokens is visualized in \textcolor{blau}{fourth column} of Fig.~\ref{fig:results-examples}, where all corners show a slightly increased similarity towards the reference token at the upper left corner of the second image, independent of their image content.

\paragraph{Quantitative results.}
To quantify our observation, we compare similarities between tokens, when they are at the border/corners and when they are not. To preclude confounding by image content, we create two versions of the representations for all images: (1) we center-crop the image before embedding, such that we arrive at a representation of shape $(w-2,h-2,c)$; and (2) we embed the original image and then discard the outermost tokens, arriving at the same shape. This allows us to compare the tokens representing the same image regions but where the outermost tokens (1) do and (2) do not lie on the border of the image during extraction. 
The \textcolor{blau}{second row} of Fig.~\ref{fig:results-quantitative} shows the average cosine similarity between all border tokens and all corner tokens for both representations. In particular, the similarities for tokens at the image border/corners during extraction are shown relative to the baseline, when they are not at the border during extraction. Relative similarity among corner tokens is increased across multiple layers, while the results for border tokens are inconclusive.

\paragraph{Implications.} The existence of corner artifacts may negatively affect dense prediction tasks that are based on similarity, such as dense correspondence. Similar to position embeddings, similarity caused by corner artifacts may obfuscate semantic (dis)similarity of image content at these locations.

\subsection{SD U-Net Representations Contain High-Norm Artifacts}\label{sec:hi-norm}
\paragraph{Qualitative observation.} We identify anomalies, which consist of groups of neighboring tokens with high norm, and high mutual similarity.
Several such anomalies can be seen in the \textcolor{gruen}{last two columns} of  Fig.~\ref{fig:results-examples}.
They primarily consist of $2\times2$ token patches that have increased Euclidean norm and high mutual cosine similarity.

\paragraph{Quantitative results.}
To analyze high-norm anomalies, we manually label their occurrences in the $L_2$ norm maps of the \texttt{up1}-block of all images in the datasets. We find that for SD-1.5, about 25\% of the images contain at least one such anomaly. In the \textcolor{gruen}{bottom row} of Fig.~\ref{fig:results-quantitative}, it can be seen that the tokens at the location of the labeled anomalies have significantly higher norm in the layers of the \texttt{up1} and \texttt{up2} blocks than the average of the tokens in the respective representations. We find the locations of the anomalies to be consistent across different layers for the same image, but not across different images, time steps, or models.

\paragraph{Implications.} Similar to the border artifacts in Sec.~\ref{sec:border-art}, high-norm anomalies may negatively affect dense prediction tasks. This includes tasks that are not similarity-based, such as depth estimation. Moreover, the observed anomalies affect the \texttt{up1} layer, commonly used for downstream tasks, and are not exclusively located at the image borders, thus potentially interfering with the representations of centered objects.

\section{Conclusion}
In this work, we presented idiosyncrasies of U-Net-based latent diffusion model representations that may provide challenges when using these representations for downstream tasks. 
We reported that these representations contain (1) a linearly extractable position embedding, (2) corner tokens of abnormally high similarity, and (3) high-norm anomalies in the up-sampling blocks. All findings are supported by both qualitative examples and quantitative analysis.

Future work may evaluate the concrete impact of these phenomena on large-scale and real-world applications. Furthermore, the causes of these phenomena, as well as their role in the generative process, should be established. For example, corner artifacts may be part of the position embedding mechanism, and high-norm tokens may function as register-like storage of global information~\citep{darcet2024}. 

\bibliography{iclr2025_conference}
\bibliographystyle{iclr2025_delta}

\appendix
\FloatBarrier

\section{Related Work}\label{appx:rw}
Initially inspired by the physical process of diffusion, diffusion models iteratively transform a distribution of noise into a desired target distribution through a sequence of learned reverse steps \cite{pmlr-v37-sohl-dickstein15,ho2020denoising}. Building on this, SD is a series of latent diffusion models for image generation \cite{rombach2022HiResSynth}, most of which employ a U-Net \cite{ronneberger2015unet} in the latent space of a pretrained variational autoencoder. More recently, transformer-based models have been added to the series~\citep{esser2024scaling}.

\paragraph{Diffusion Models for Representation Learning.}
Diffusion models, and SD in particular, have been analyzed and used for representation learning as a basis for a variety of downstream tasks. While some works modify the model architecture or training process specifically for representation learning \citep{hudson2023soda,chen2024deconstructingdenoisingdiffusionmodels}, many works use the intermediate representations of pretrained models. Common SD versions used in the literature are SD-1.5 and SD-2.1 \citep{luo2023dhf,zhao2023unleashing,zhang2023tale,zhang2024telling,stracke2024clean,linhardt2024analysis}.
Various works have investigated different aspects of the learned representations, finding that semantic information is captured in the bottleneck layers of the U-Net \citep{kwon2023diffusion,park2023unsupervised}. 
Other works studied image diffusion models' alignment to human representations and human-like shape bias \citep{linhardt2024analysis,jaini2024intriguing}.

\paragraph{SD Representations for Downstream Tasks.}
In recent years, there has been substantial interest in exploring the suitability of SD representations for downstream tasks, such as classification \citep{xiang2023denoising,mukhopadhyay2023diffusion,stracke2024clean}, semantic correspondence \citep{zhang2023tale,zhang2024telling,banani2024probing,tang2023emergent,luo2023dhf,hedlin2023unsupervised,li2023sd4match,stracke2024clean,fundel2024distillationdiffusionfeaturessemantic,mariotti2024improving,kim2025matchme}, semantic segmentation \citep{baranchuk2022labelefficient,zhao2023unleashing,ji2024diffusion,couairon2024zeroshot,tian2024diffuse,zhang2025three}, and depth estimation \citep{Chen2023BeyondSS,zhao2023unleashing,Patni2024ECoDepth,stracke2024clean,zhang2025three}.
It has been observed that downstream task performance tends to increase with the number of pre-training iterations \citep{zhao2023unleashing, zhang2025three}. 
Multiple works reported that \texttt{up}-blocks of the U-Net contain the most useful representations for downstream tasks \citep{zhang2023tale,banani2024probing,stracke2024clean}. \cite{tang2023emergent} suggest that \texttt{up}-blocks lower in the U-Net yield more semantically-aware representations, while \texttt{up}-blocks higher in the U-Net focus more on more low-level details.

\section{Results for SD-2.1 and SD-Turbo}\label{appx:models}

\begin{figure}
\includegraphics[width=1\textwidth]{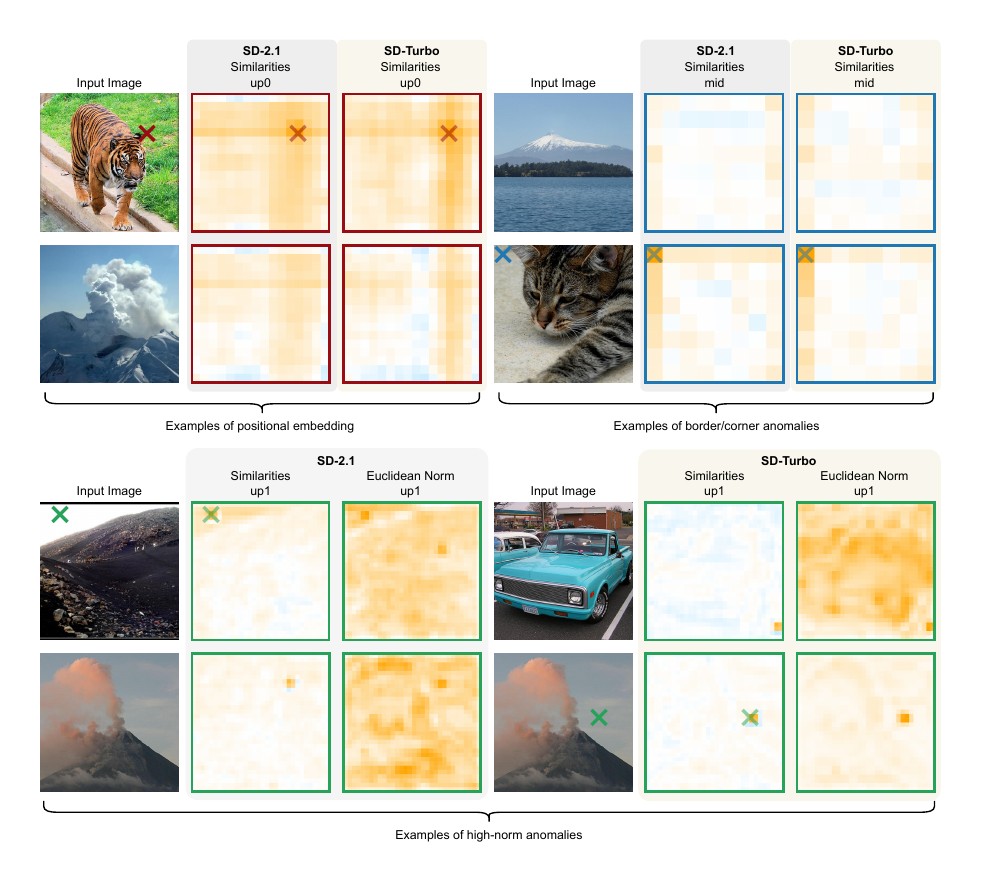}
\caption{Cosine similarity and Euclidean norm for representations of SD-2.1 and SD-Turbo. The similarities are relative to the representation token at the image and location of the marker in the respective image pair. \textbf{Top left:} Positional embedding for SD-2.1 (left), and SD-Turbo (right). \textbf{Top right:} Corner/border anomalies for SD-2.1 (left), and SD-Turbo (right). \textbf{Bottom:} High-norm anomalies for SD-2.1 (left), and SD-Turbo (right).}
\label{fig:results-examples-appendix}
\end{figure}

\begin{figure}
\includegraphics[width=1\textwidth]{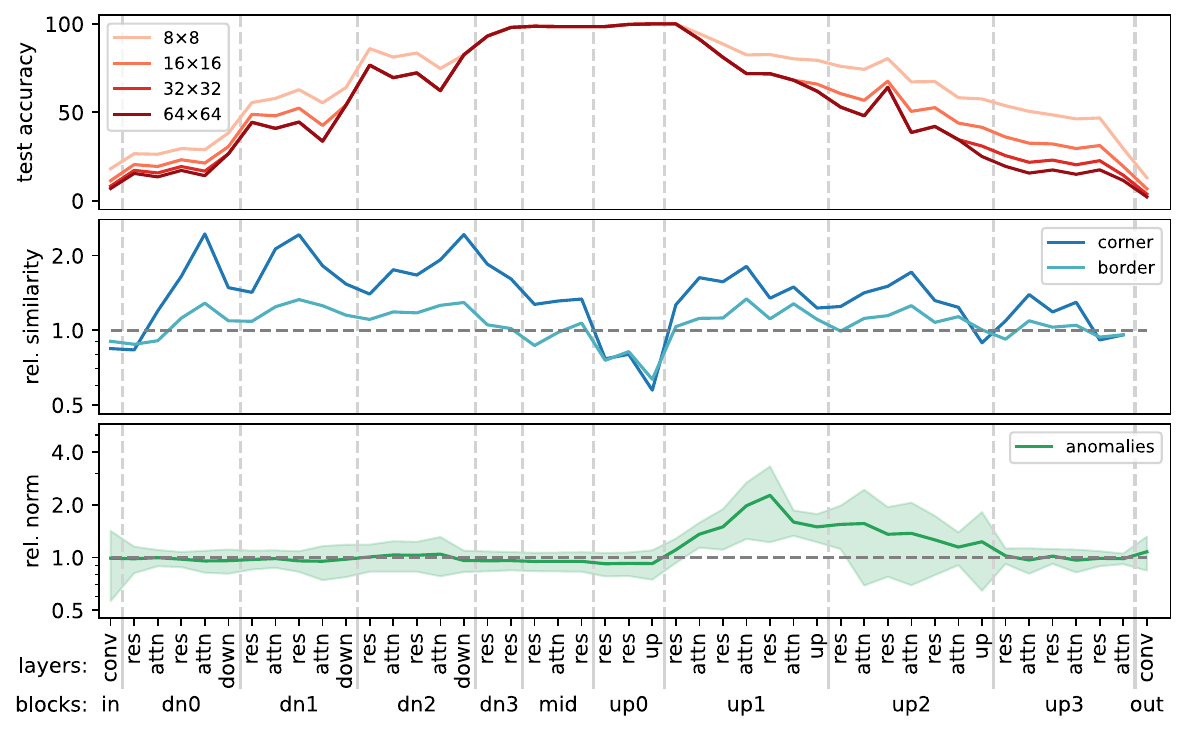}
\caption{Quantitative results for SD-2.1. \textcolor{rot}{\textbf{Top row:}} Linear probe accuracy for position estimation. Brighter shades indicate reduced resolution. \textcolor{blau}{\textbf{Middle row:}} Relative similarity of tokens lying at a border/corner of the cropped images w.r.t. their similarity before cropping. (log-2 scale). \textcolor{gruen}{\textbf{Bottom row:}} Relative average norm of anomalous tokens w.r.t. to the mean norm of all tokens of the same representation (log-2 scale).}
\label{fig:results-quantitative-sd21}
\end{figure}

\begin{figure}
\includegraphics[width=1\textwidth]{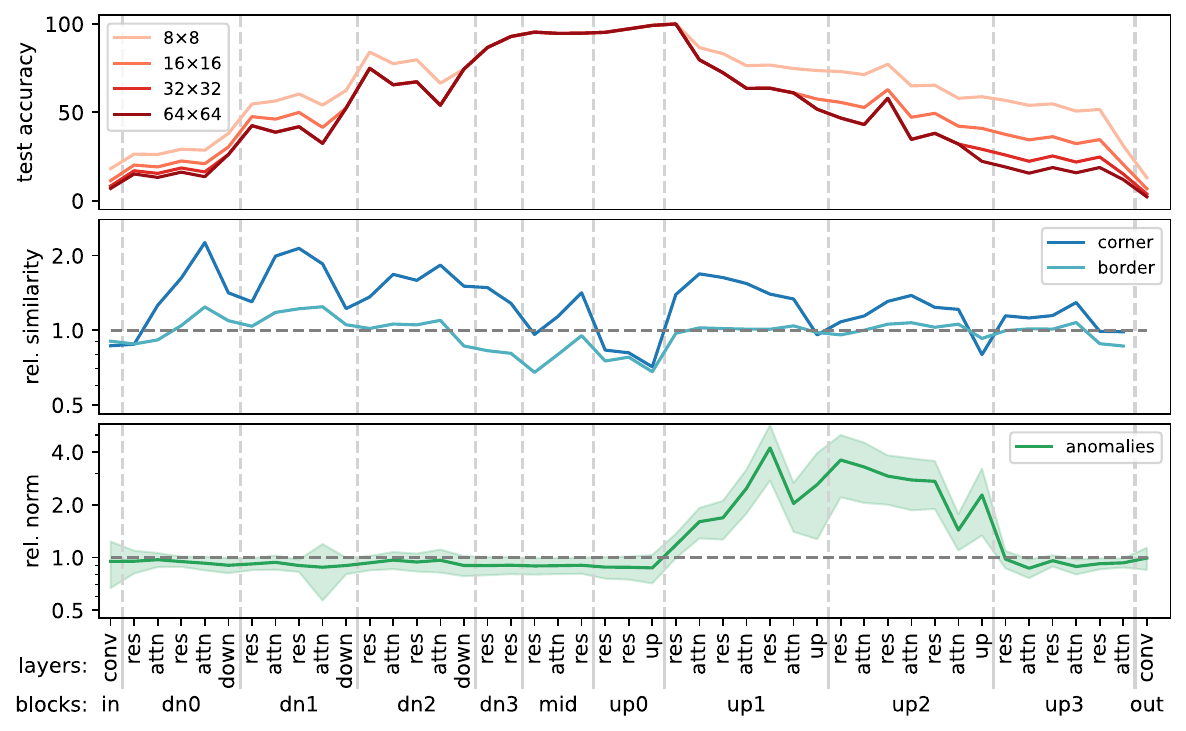}
\caption{Quantitative results for SD-Turbo. \textcolor{rot}{\textbf{Top row:}} Linear probe accuracy for position estimation. Brighter shades indicate reduced resolution. \textcolor{blau}{\textbf{Middle row:}} Relative similarity of tokens lying at a border/corner of the cropped images w.r.t. their similarity before cropping. (log-2 scale). \textcolor{gruen}{\textbf{Bottom row:}} Relative average norm of anomalous tokens w.r.t. to the mean norm of all tokens of the same representation (log-2 scale).}
\label{fig:results-quantitative-sdturbo}
\end{figure}

Complementary to the results on SD-1.5 presented in the main text, we here provide results for SD-2.1 and SD-Turbo, which are based on the same model architecture \citep{rombach2022HiResSynth}. Fig.~\ref{fig:results-examples-appendix} shows additional examples for the three phenomena described in the main text. Fig.~\ref{fig:results-quantitative-sd21} shows the results for the quantitative experiments on SD-2.1, and Fig.~\ref{fig:results-quantitative-sdturbo} on SD-Turbo. The results are overall consistent across all evaluated models, suggesting that our findings are not limited to a specific model. 

\end{document}

%% file: math_commands.tex
\usepackage{amsmath,amsfonts,bm}

\def\eqref#1{equation~\ref{#1}}

\def\1{\bm{1}}

\DeclareMathAlphabet{\mathsfit}{\encodingdefault}{\sfdefault}{m}{sl}
\SetMathAlphabet{\mathsfit}{bold}{\encodingdefault}{\sfdefault}{bx}{n}













%% file: iclr2025_conference.bbl
\begin{thebibliography}{47}
\providecommand{\natexlab}[1]{#1}
\providecommand{\url}[1]{\texttt{#1}}
\expandafter\ifx\csname urlstyle\endcsname\relax
  \providecommand{\doi}[1]{doi: #1}\else
  \providecommand{\doi}{doi: \begingroup \urlstyle{rm}\Url}\fi

\bibitem[Baranchuk et~al.(2022)Baranchuk, Voynov, Rubachev, Khrulkov, and
  Babenko]{baranchuk2022labelefficient}
Dmitry Baranchuk, Andrey Voynov, Ivan Rubachev, Valentin Khrulkov, and Artem
  Babenko.
\newblock Label-efficient semantic segmentation with diffusion models.
\newblock In \emph{Proceedings of the International Conference on Learning
  Representations}, 2022.
\newblock URL \url{https://openreview.net/forum?id=SlxSY2UZQT}.

\bibitem[Caron et~al.(2021)Caron, Touvron, Misra, J\'egou, Mairal, Bojanowski,
  and Joulin]{dino}
Mathilde Caron, Hugo Touvron, Ishan Misra, Herv\'e J\'egou, Julien Mairal,
  Piotr Bojanowski, and Armand Joulin.
\newblock Emerging properties in self-supervised vision transformers.
\newblock In \emph{Proceedings of the IEEE/CVF International Conference on
  Computer Vision (ICCV)}, pp.\  9650--9660, October 2021.

\bibitem[Chen et~al.(2024)Chen, Liu, Xie, and
  He]{chen2024deconstructingdenoisingdiffusionmodels}
Xinlei Chen, Zhuang Liu, Saining Xie, and Kaiming He.
\newblock Deconstructing denoising diffusion models for self-supervised
  learning.
\newblock \emph{arXiv preprint arXiv:2401.14404}, 2024.
\newblock URL \url{https://arxiv.org/abs/2401.14404}.

\bibitem[Chen et~al.(2023)Chen, Vi'egas, and Wattenberg]{Chen2023BeyondSS}
Yida Chen, Fernanda Vi'egas, and Martin Wattenberg.
\newblock Beyond surface statistics: Scene representations in a latent
  diffusion model.
\newblock \emph{arXiv preprint arXiv:2306.05720}, 2023.
\newblock URL \url{https://arxiv.org/abs/2306.05720}.

\bibitem[Couairon et~al.(2024)Couairon, Shukor, Haugeard, Cord, and
  Thome]{couairon2024zeroshot}
Paul Couairon, Mustafa Shukor, Jean-Emmanuel Haugeard, Matthieu Cord, and
  Nicolas Thome.
\newblock Diffcut: Catalyzing zero-shot semantic segmentation with diffusion
  features and recursive normalized cut.
\newblock In \emph{Proceedings of the Thirty-eighth Annual Conference on Neural
  Information Processing Systems}, 2024.
\newblock URL \url{https://openreview.net/forum?id=N0xNf9Qqmc}.

\bibitem[Darcet et~al.(2024)Darcet, Oquab, Mairal, and Bojanowski]{darcet2024}
Timoth{\'e}e Darcet, Maxime Oquab, Julien Mairal, and Piotr Bojanowski.
\newblock Vision transformers need registers.
\newblock In \emph{The Twelfth International Conference on Learning
  Representations {ICLR} 2024}, 2024.
\newblock URL \url{https://openreview.net/forum?id=2dnO3LLiJ1}.

\bibitem[Dhariwal \& Nichol(2021)Dhariwal and Nichol]{dhariwal2021diffusion}
Prafulla Dhariwal and Alexander Nichol.
\newblock Diffusion models beat gans on image synthesis.
\newblock \emph{Advances in Neural Information Processing Systems},
  34:\penalty0 8780--8794, 2021.

\bibitem[El~Banani et~al.(2024)El~Banani, Raj, Maninis, Kar, Li, Rubinstein,
  Sun, Guibas, Johnson, and Jampani]{banani2024probing}
Mohamed El~Banani, Amit Raj, Kevis-Kokitsi Maninis, Abhishek Kar, Yuanzhen Li,
  Michael Rubinstein, Deqing Sun, Leonidas Guibas, Justin Johnson, and Varun
  Jampani.
\newblock Probing the 3d awareness of visual foundation models.
\newblock In \emph{Proceedings of the IEEE/CVF Conference on Computer Vision
  and Pattern Recognition (CVPR)}, pp.\  21795--21806, 2024.
\newblock URL
  \url{https://openaccess.thecvf.com/content/CVPR2024/html/Banani_Probing_the_3D_Awareness_of_Visual_Foundation_Models_CVPR_2024_paper.html}.

\bibitem[Esser et~al.(2024)Esser, Kulal, Blattmann, Entezari, M{\"u}ller,
  Saini, Levi, Lorenz, Sauer, Boesel, Podell, Dockhorn, English, and
  Rombach]{esser2024scaling}
Patrick Esser, Sumith Kulal, Andreas Blattmann, Rahim Entezari, Jonas
  M{\"u}ller, Harry Saini, Yam Levi, Dominik Lorenz, Axel Sauer, Frederic
  Boesel, Dustin Podell, Tim Dockhorn, Zion English, and Robin Rombach.
\newblock Scaling rectified flow transformers for high-resolution image
  synthesis.
\newblock In \emph{Forty-first International Conference on Machine Learning},
  2024.
\newblock URL \url{https://openreview.net/forum?id=FPnUhsQJ5B}.

\bibitem[Fundel et~al.(2024)Fundel, Schusterbauer, Hu, and
  Ommer]{fundel2024distillationdiffusionfeaturessemantic}
Frank Fundel, Johannes Schusterbauer, Vincent~Tao Hu, and Björn Ommer.
\newblock Distillation of diffusion features for semantic correspondence.
\newblock \emph{arXiv preprint arXiv:2412.03512}, 2024.
\newblock URL \url{https://arxiv.org/abs/2412.03512}.

\bibitem[Goodfellow et~al.(2020)Goodfellow, Pouget-Abadie, Mirza, Xu,
  Warde-Farley, Ozair, Courville, and Bengio]{goodfellow2020generative}
Ian Goodfellow, Jean Pouget-Abadie, Mehdi Mirza, Bing Xu, David Warde-Farley,
  Sherjil Ozair, Aaron Courville, and Yoshua Bengio.
\newblock Generative adversarial networks.
\newblock \emph{Communications of the ACM}, 63\penalty0 (11):\penalty0
  139--144, 2020.

\bibitem[Hedlin et~al.(2023)Hedlin, Sharma, Mahajan, Isack, Kar, Tagliasacchi,
  and Yi]{hedlin2023unsupervised}
Eric Hedlin, Gopal Sharma, Shweta Mahajan, Hossam Isack, Abhishek Kar, Andrea
  Tagliasacchi, and Kwang~Moo Yi.
\newblock Unsupervised semantic correspondence using stable diffusion.
\newblock \emph{Advances in Neural Information Processing Systems},
  36:\penalty0 8266--8279, 2023.
\newblock URL
  \url{https://proceedings.neurips.cc/paper_files/paper/2023/file/1a074a28c3a6f2056562d00649ae6416-Paper-Conference.pdf}.

\bibitem[Ho et~al.(2020)Ho, Jain, and Abbeel]{ho2020denoising}
Jonathan Ho, Ajay Jain, and Pieter Abbeel.
\newblock Denoising diffusion probabilistic models.
\newblock \emph{Advances in Neural Information Processing Systems},
  33:\penalty0 6840--6851, 2020.

\bibitem[Hudson et~al.(2024)Hudson, Zoran, Malinowski, Lampinen, Jaegle,
  McClelland, Matthey, Hill, and Lerchner]{hudson2023soda}
Drew~A. Hudson, Daniel Zoran, Mateusz Malinowski, Andrew~K. Lampinen, Andrew
  Jaegle, James~L. McClelland, Loic Matthey, Felix Hill, and Alexander
  Lerchner.
\newblock Soda: Bottleneck diffusion models for representation learning.
\newblock In \emph{Proceedings of the IEEE/CVF Conference on Computer Vision
  and Pattern Recognition (CVPR)}, pp.\  23115--23127, 2024.

\bibitem[Jaini et~al.(2024)Jaini, Clark, and Geirhos]{jaini2024intriguing}
Priyank Jaini, Kevin Clark, and Robert Geirhos.
\newblock Intriguing properties of generative classifiers.
\newblock In \emph{Proceedings of the 12th International Conference on Learning
  Representations ({ICLR})}, 2024.
\newblock URL \url{https://openreview.net/forum?id=rmg0qMKYRQ}.

\bibitem[Ji et~al.(2024)Ji, He, Qu, Tan, Qin, and Wu]{ji2024diffusion}
Yuxiang Ji, Boyong He, Chenyuan Qu, Zhuoyue Tan, Chuan Qin, and Liaoni Wu.
\newblock Diffusion features to bridge domain gap for semantic segmentation.
\newblock \emph{arXiv preprint arXiv:2406.00777}, 2024.
\newblock URL \url{https://arxiv.org/abs/2406.00777}.

\bibitem[Ke et~al.(2024)Ke, Obukhov, Huang, Metzger, Daudt, and
  Schindler]{Ke2024RepurposingDI}
Bingxin Ke, Anton Obukhov, Shengyu Huang, Nando Metzger, Rodrigo~Caye Daudt,
  and Konrad Schindler.
\newblock Repurposing diffusion-based image generators for monocular depth
  estimation.
\newblock In \emph{2024 IEEE/CVF Conference on Computer Vision and Pattern
  Recognition (CVPR)}, pp.\  9492--9502, 2024.
\newblock \doi{10.1109/CVPR52733.2024.00907}.

\bibitem[Kim et~al.(2025)Kim, Heo, Yun, Kim, and Han]{kim2025matchme}
Jiwon Kim, Byeongho Heo, Sangdoo Yun, Seungryong Kim, and Dongyoon Han.
\newblock Match me if you can: Semi-supervised semantic correspondence learning
  with unpaired images.
\newblock In \emph{Computer Vision -- ACCV 2024}, pp.\  462--479. Springer
  Nature Singapore, 2025.
\newblock URL
  \url{https://link.springer.com/chapter/10.1007/978-981-96-0960-4_28}.

\bibitem[Kingma \& Ba(2017)Kingma and Ba]{kingma2017adam}
Diederik~P. Kingma and Jimmy Ba.
\newblock Adam: A method for stochastic optimization.
\newblock \emph{arXiv preprint arXiv:1412.6980}, 2017.
\newblock URL \url{https://arxiv.org/abs/1412.6980}.

\bibitem[Kwon et~al.(2023)Kwon, Jeong, and Uh]{kwon2023diffusion}
Mingi Kwon, Jaeseok Jeong, and Youngjung Uh.
\newblock Diffusion models already have a semantic latent space.
\newblock In \emph{The Eleventh International Conference on Learning
  Representations}, 2023.
\newblock URL \url{https://openreview.net/forum?id=pd1P2eUBVfq}.

\bibitem[Li et~al.(2024)Li, Lu, Han, and Prisacariu]{li2023sd4match}
Xinghui Li, Jingyi Lu, Kai Han, and Victor~Adrian Prisacariu.
\newblock Sd4match: Learning to prompt stable diffusion model for semantic
  matching.
\newblock In \emph{Proceedings of the IEEE/CVF Conference on Computer Vision
  and Pattern Recognition (CVPR)}, pp.\  27558--27568, 2024.
\newblock URL
  \url{https://openaccess.thecvf.com/content/CVPR2024/html/Li_SD4Match_Learning_to_Prompt_Stable_Diffusion_Model_for_Semantic_Matching_CVPR_2024_paper.html}.

\bibitem[Linhardt et~al.(2024)Linhardt, Morik, Bender, and
  Borras]{linhardt2024analysis}
Lorenz Linhardt, Marco Morik, Sidney Bender, and Naima~Elosegui Borras.
\newblock An analysis of human alignment of latent diffusion models.
\newblock In \emph{ICLR 2024 Workshop on Representational Alignment}, 2024.
\newblock URL \url{https://openreview.net/forum?id=PFnoxcKh33}.

\bibitem[Luo et~al.(2023)Luo, Dunlap, Park, Holynski, and Darrell]{luo2023dhf}
Grace Luo, Lisa Dunlap, Dong~Huk Park, Aleksander Holynski, and Trevor Darrell.
\newblock Diffusion hyperfeatures: Searching through time and space for
  semantic correspondence.
\newblock In \emph{Advances in Neural Information Processing Systems}, 2023.
\newblock URL \url{https://openreview.net/forum?id=Vm1zeYqwdc}.

\bibitem[Mariotti et~al.(2024)Mariotti, Aodha, and
  Bilen]{mariotti2024improving}
Octave Mariotti, Oisin~Mac Aodha, and Hakan Bilen.
\newblock Improving semantic correspondence with viewpoint-guided spherical
  maps.
\newblock In \emph{2024 IEEE/CVF Conference on Computer Vision and Pattern
  Recognition (CVPR)}, pp.\  19521--19530, 2024.
\newblock \doi{10.1109/CVPR52733.2024.01846}.

\bibitem[Mukhopadhyay et~al.(2023)Mukhopadhyay, Gwilliam, Agarwal, Padmanabhan,
  Swaminathan, Hegde, Zhou, and Shrivastava]{mukhopadhyay2023diffusion}
Soumik Mukhopadhyay, Matthew Gwilliam, Vatsal Agarwal, Namitha Padmanabhan,
  Archana Swaminathan, Srinidhi Hegde, Tianyi Zhou, and Abhinav Shrivastava.
\newblock Diffusion models beat gans on image classification.
\newblock \emph{arXiv preprint arXiv:2307.08702}, 2023.
\newblock URL \url{https://arxiv.org/abs/2307.08702}.

\bibitem[Oquab et~al.(2024)Oquab, Darcet, Moutakanni, Vo, Szafraniec, Khalidov,
  Fernandez, HAZIZA, Massa, El-Nouby, Assran, Ballas, Galuba, Howes, Huang, Li,
  Misra, Rabbat, Sharma, Synnaeve, Xu, Jegou, Mairal, Labatut, Joulin, and
  Bojanowski]{dinov2}
Maxime Oquab, Timoth{\'e}e Darcet, Th{\'e}o Moutakanni, Huy~V. Vo, Marc
  Szafraniec, Vasil Khalidov, Pierre Fernandez, Daniel HAZIZA, Francisco Massa,
  Alaaeldin El-Nouby, Mido Assran, Nicolas Ballas, Wojciech Galuba, Russell
  Howes, Po-Yao Huang, Shang-Wen Li, Ishan Misra, Michael Rabbat, Vasu Sharma,
  Gabriel Synnaeve, Hu~Xu, Herve Jegou, Julien Mairal, Patrick Labatut, Armand
  Joulin, and Piotr Bojanowski.
\newblock {DINO}v2: Learning robust visual features without supervision.
\newblock \emph{Transactions on Machine Learning Research}, 2024.
\newblock ISSN 2835-8856.
\newblock URL \url{https://openreview.net/forum?id=a68SUt6zFt}.

\bibitem[Park et~al.(2023)Park, Kwon, Jo, and Uh]{park2023unsupervised}
Yong-Hyun Park, Mingi Kwon, Junghyo Jo, and Youngjung Uh.
\newblock Unsupervised discovery of semantic latent directions in diffusion
  models.
\newblock \emph{arXiv preprint arXiv:2302.12469}, 2023.
\newblock URL \url{https://arxiv.org/abs/2302.12469}.

\bibitem[Patni et~al.(2024)Patni, Agarwal, and Arora]{Patni2024ECoDepth}
Suraj Patni, Aradhye Agarwal, and Chetan Arora.
\newblock Ecodepth: Effective conditioning of diffusion models for monocular
  depth estimation.
\newblock In \emph{2024 IEEE/CVF Conference on Computer Vision and Pattern
  Recognition (CVPR)}, pp.\  28285--28295, 2024.
\newblock \doi{10.1109/CVPR52733.2024.02672}.

\bibitem[Peebles \& Xie(2023)Peebles and Xie]{peebles2023scalable}
William Peebles and Saining Xie.
\newblock Scalable diffusion models with transformers.
\newblock In \emph{2023 IEEE/CVF International Conference on Computer Vision
  (ICCV)}, pp.\  4172--4182, 2023.
\newblock \doi{10.1109/ICCV51070.2023.00387}.

\bibitem[Radford et~al.(2021)Radford, Kim, Hallacy, Ramesh, Goh, Agarwal,
  Sastry, Askell, Mishkin, Clark, Krueger, and Sutskever]{clip}
Alec Radford, Jong~Wook Kim, Chris Hallacy, Aditya Ramesh, Gabriel Goh,
  Sandhini Agarwal, Girish Sastry, Amanda Askell, Pamela Mishkin, Jack Clark,
  Gretchen Krueger, and Ilya Sutskever.
\newblock Learning transferable visual models from natural language
  supervision.
\newblock In \emph{Proceedings of the 38th International Conference on Machine
  Learning}, volume 139 of \emph{Proceedings of Machine Learning Research},
  pp.\  8748--8763. PMLR, 18--24 Jul 2021.
\newblock URL \url{https://proceedings.mlr.press/v139/radford21a.html}.

\bibitem[Rassin et~al.(2023)Rassin, Hirsch, Glickman, Ravfogel, Goldberg, and
  Chechik]{rassin2023}
Royi Rassin, Eran Hirsch, Daniel Glickman, Shauli Ravfogel, Yoav Goldberg, and
  Gal Chechik.
\newblock Linguistic binding in diffusion models: Enhancing attribute
  correspondence through attention map alignment.
\newblock In A.~Oh, T.~Naumann, A.~Globerson, K.~Saenko, M.~Hardt, and
  S.~Levine (eds.), \emph{Advances in Neural Information Processing Systems},
  volume~36, pp.\  3536--3559. Curran Associates, Inc., 2023.
\newblock URL
  \url{https://proceedings.neurips.cc/paper_files/paper/2023/file/0b08d733a5d45a547344c4e9d88bb8bc-Paper-Conference.pdf}.

\bibitem[Rombach et~al.(2022)Rombach, Blattmann, Lorenz, Esser, and
  Ommer]{rombach2022HiResSynth}
Robin Rombach, Andreas Blattmann, Dominik Lorenz, Patrick Esser, and Bj\"orn
  Ommer.
\newblock High-resolution image synthesis with latent diffusion models.
\newblock In \emph{Proceedings of the IEEE/CVF Conference on Computer Vision
  and Pattern Recognition (CVPR)}, pp.\  10684--10695, 2022.

\bibitem[Ronneberger et~al.(2015)Ronneberger, Fischer, and
  Brox]{ronneberger2015unet}
Olaf Ronneberger, Philipp Fischer, and Thomas Brox.
\newblock U-net: Convolutional networks for biomedical image segmentation.
\newblock In \emph{Medical Image Computing and Computer-Assisted Intervention
  -- MICCAI 2015}, pp.\  234--241. Springer International Publishing, 2015.
\newblock ISBN 978-3-319-24574-4.
\newblock \doi{10.1007/978-3-319-24574-4_28}.

\bibitem[Russakovsky et~al.(2015)Russakovsky, Deng, Su, Krause, Satheesh, Ma,
  Huang, Karpathy, Khosla, Bernstein, Berg, and Fei-Fei]{ILSVRC15}
Olga Russakovsky, Jia Deng, Hao Su, Jonathan Krause, Sanjeev Satheesh, Sean Ma,
  Zhiheng Huang, Andrej Karpathy, Aditya Khosla, Michael Bernstein,
  Alexander~C. Berg, and Li~Fei-Fei.
\newblock {ImageNet Large Scale Visual Recognition Challenge}.
\newblock \emph{International Journal of Computer Vision (IJCV)}, 115:\penalty0
  211--252, 2015.
\newblock \doi{10.1007/s11263-015-0816-y}.

\bibitem[Sauer et~al.(2024)Sauer, Boesel, Dockhorn, Blattmann, Esser, and
  Rombach]{sauer2024fast}
Axel Sauer, Frederic Boesel, Tim Dockhorn, Andreas Blattmann, Patrick Esser,
  and Robin Rombach.
\newblock Fast high-resolution image synthesis with latent adversarial
  diffusion distillation.
\newblock In \emph{SIGGRAPH Asia 2024 Conference Papers}, SA '24. Association
  for Computing Machinery, 2024.
\newblock ISBN 9798400711312.
\newblock \doi{10.1145/3680528.3687625}.

\bibitem[Sauer et~al.(2025)Sauer, Lorenz, Blattmann, and
  Rombach]{sauer2023adversarial}
Axel Sauer, Dominik Lorenz, Andreas Blattmann, and Robin Rombach.
\newblock Adversarial diffusion distillation.
\newblock In Ale{\v{s}} Leonardis, Elisa Ricci, Stefan Roth, Olga Russakovsky,
  Torsten Sattler, and G{\"u}l Varol (eds.), \emph{Computer Vision -- ECCV
  2024}, pp.\  87--103, Cham, 2025. Springer Nature Switzerland.
\newblock ISBN 978-3-031-73016-0.

\bibitem[Sohl-Dickstein et~al.(2015{\natexlab{a}})Sohl-Dickstein, Weiss,
  Maheswaranathan, and Ganguli]{pmlr-v37-sohl-dickstein15}
Jascha Sohl-Dickstein, Eric Weiss, Niru Maheswaranathan, and Surya Ganguli.
\newblock Deep unsupervised learning using nonequilibrium thermodynamics.
\newblock In \emph{Proceedings of the 32nd International Conference on Machine
  Learning}, volume~37 of \emph{Proceedings of Machine Learning Research}, pp.\
   2256--2265. PMLR, 2015{\natexlab{a}}.
\newblock URL \url{https://proceedings.mlr.press/v37/sohl-dickstein15.html}.

\bibitem[Sohl-Dickstein et~al.(2015{\natexlab{b}})Sohl-Dickstein, Weiss,
  Maheswaranathan, and Ganguli]{sohl2015deep}
Jascha Sohl-Dickstein, Eric Weiss, Niru Maheswaranathan, and Surya Ganguli.
\newblock Deep unsupervised learning using nonequilibrium thermodynamics.
\newblock In \emph{International Conference on Machine Learning}, pp.\
  2256--2265. PMLR, 2015{\natexlab{b}}.

\bibitem[Song \& Ermon(2019)Song and Ermon]{song2019}
Yang Song and Stefano Ermon.
\newblock Generative modeling by estimating gradients of the data distribution.
\newblock In \emph{Proceedings of the 33rd International Conference on Neural
  Information Processing Systems}, Red Hook, NY, USA, 2019. Curran Associates
  Inc.

\bibitem[Stracke et~al.(2024)Stracke, Baumann, Bauer, Fundel, and
  Ommer]{stracke2024clean}
Nick Stracke, Stefan~Andreas Baumann, Kolja Bauer, Frank Fundel, and Björn
  Ommer.
\newblock Cleandift: Diffusion features without noise.
\newblock \emph{arXiv preprint arXiv:2412.03439}, 2024.
\newblock URL \url{https://arxiv.org/abs/2412.03439}.

\bibitem[Tang et~al.(2023)Tang, Jia, Wang, Phoo, and
  Hariharan]{tang2023emergent}
Luming Tang, Menglin Jia, Qianqian Wang, Cheng~Perng Phoo, and Bharath
  Hariharan.
\newblock Emergent correspondence from image diffusion.
\newblock In \emph{Thirty-seventh Conference on Neural Information Processing
  Systems}, 2023.
\newblock URL \url{https://openreview.net/forum?id=ypOiXjdfnU}.

\bibitem[Tian et~al.(2024)Tian, Aggarwal, Colaco, Kira, and
  Gonzalez-Franco]{tian2024diffuse}
Junjiao Tian, Lavisha Aggarwal, Andrea Colaco, Zsolt Kira, and Mar
  Gonzalez-Franco.
\newblock {Diffuse, Attend, and Segment: Unsupervised Zero-Shot Segmentation
  using Stable Diffusion}.
\newblock In \emph{2024 IEEE/CVF Conference on Computer Vision and Pattern
  Recognition (CVPR)}, pp.\  3554--3563. IEEE Computer Society, 2024.
\newblock \doi{10.1109/CVPR52733.2024.00341}.

\bibitem[Xiang et~al.(2023)Xiang, Yang, Huang, and Wang]{xiang2023denoising}
Weilai Xiang, Hongyu Yang, Di~Huang, and Yunhong Wang.
\newblock Denoising diffusion autoencoders are unified self-supervised
  learners.
\newblock In \emph{Proceedings of the IEEE/CVF International Conference on
  Computer Vision (ICCV)}, pp.\  15802--15812, 2023.
\newblock \doi{10.1109/ICCV51070.2023.01448}.

\bibitem[Zhang et~al.(2023)Zhang, Herrmann, Hur, Cabrera, Jampani, Sun, and
  Yang]{zhang2023tale}
Junyi Zhang, Charles Herrmann, Junhwa Hur, Luisa~Polania Cabrera, Varun
  Jampani, Deqing Sun, and Ming-Hsuan Yang.
\newblock A tale of two features: Stable diffusion complements dino for
  zero-shot semantic correspondence.
\newblock \emph{Advances in Neural Information Processing Systems},
  36:\penalty0 45533--45547, 2023.
\newblock URL
  \url{https://proceedings.neurips.cc/paper_files/paper/2023/file/8e9bdc23f169a05ea9b72ccef4574551-Paper-Conference.pdf}.

\bibitem[Zhang et~al.(2024)Zhang, Herrmann, Hur, Chen, Jampani, Sun, and
  Yang]{zhang2024telling}
Junyi Zhang, Charles Herrmann, Junhwa Hur, Eric Chen, Varun Jampani, Deqing
  Sun, and Ming-Hsuan Yang.
\newblock Telling left from right: Identifying geometry-aware semantic
  correspondence.
\newblock In \emph{Proceedings of the IEEE/CVF Conference on Computer Vision
  and Pattern Recognition (CVPR)}, pp.\  3076--3085, June 2024.

\bibitem[Zhang et~al.(2025)Zhang, Song, Shi, Liu, and Li]{zhang2025three}
Manyuan Zhang, Guanglu Song, Xiaoyu Shi, Yu~Liu, and Hongsheng Li.
\newblock Three things we need to know about transferring stable diffusion to
  visual dense prediction tasks.
\newblock In \emph{Computer Vision -- ECCV 2024}, pp.\  128--145. Springer
  Nature Switzerland, 2025.
\newblock ISBN 978-3-031-72946-1.
\newblock URL
  \url{https://www.ecva.net/papers/eccv_2024/papers_ECCV/papers/05837.pdf}.

\bibitem[Zhao et~al.(2023)Zhao, Rao, Liu, Liu, Zhou, and
  Lu]{zhao2023unleashing}
Wenliang Zhao, Yongming Rao, Zuyan Liu, Benlin Liu, Jie Zhou, and Jiwen Lu.
\newblock Unleashing text-to-image diffusion models for visual perception.
\newblock In \emph{2023 IEEE/CVF International Conference on Computer Vision
  (ICCV)}, pp.\  5706--5716. IEEE Computer Society, 2023.
\newblock \doi{10.1109/ICCV51070.2023.00527}.

\end{thebibliography}
